\pdfoutput=1

\documentclass[11pt]{article}

\usepackage[preprint]{acl}

\usepackage{times}
\usepackage{latexsym}

\usepackage[T1]{fontenc}

\usepackage[utf8]{inputenc}

\usepackage{microtype}

\usepackage{inconsolata}

\usepackage{graphicx}
\usepackage{graphicx}
\usepackage{booktabs}
\usepackage{multicol}
\usepackage{multirow}
\usepackage{colortbl}
\usepackage{arydshln}
\usepackage{makecell}
\usepackage{stfloats}
\usepackage{url}
\usepackage{stfloats}
\usepackage{xspace}
\usepackage{balance}

\newcommand{\method}{\textsc{SENSE}\xspace}

\title{Rethinking Semantic Parsing for Large Language Models: Enhancing LLM Performance with Semantic Hints}

\author{Kaikai An\textsuperscript{1,2}\thanks{~~Equal contribution}, Shuzheng Si\textsuperscript{1,2$*$}, Helan Hu\textsuperscript{1,2},  Haozhe Zhao\textsuperscript{1,2}, \\
\textbf{Yuchi Wang\textsuperscript{1},}  \textbf{Qingyan Guo\textsuperscript{3},}  \textbf{Baobao Chang\textsuperscript{1}\thanks{~~Corresponding author}} \\
\textsuperscript{1} National Key Laboratory for Multimedia Information Processing, Peking University \\
\textsuperscript{2} School of Software and Microelectronics, Peking University     \textsuperscript{3} Tsinghua University \\
\texttt{ankaikai@stu.pku.edu.cn, chbb@pku.edu.cn}
}

\begin{document}
\maketitle
\begin{abstract}
Semantic Parsing aims to capture the meaning of a sentence and convert it into a logical, structured form. Previous studies show that semantic parsing enhances the performance of smaller models (e.g., BERT) on downstream tasks. However, it remains unclear whether the improvements extend similarly to LLMs. In this paper, our empirical findings reveal that, unlike smaller models, directly adding semantic parsing results into LLMs reduces their performance. To overcome this, we propose \method, a novel prompting approach that embeds semantic hints within the prompt. Experiments show that \method\ consistently improves LLMs' performance across various tasks, highlighting the potential of integrating semantic information to improve LLM capabilities.
\end{abstract}

\section{Introduction}
Semantic Parsing is a fundamental task in Natural Language Processing, which involves converting a natural language sentence into structured meaning representation. This includes tasks like Semantic Role Labeling (SRL), Frame Semantic Parsing (FSP) and Abstract Meaning Representation (AMR) \cite{gildea2002automatic,baker-etal-2007-semeval, Banarescu_Bonial_Cai_Georgescu_Griffitt_Hermjakob_Knight_Koehn_Palmer_Schneider_2013, Palmer_Titov_Wu_2010, an-etal-2023-coarse}. 
Such structured information are applicable across various tasks, like Question Answering \cite{Khashabi_Khot_Sabharwal_Roth_2022}, Machine Translation \cite{rapp2022using}, Dialogue Systems \cite{xu2020semantic, si-etal-2022-mining, si2024spokenwoz} and so on.

\begin{figure}[t]
    \centering
    \includegraphics[width=0.8\linewidth]{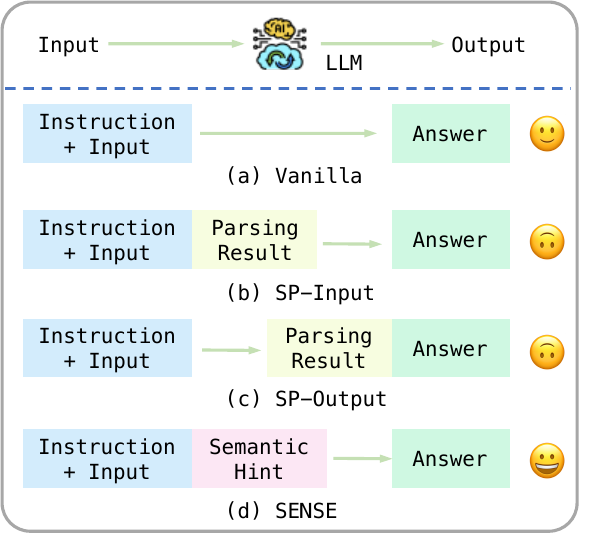}
    \caption{Different ways of evaluating LLMs on downstream tasks. While (a) represents direct prompting models, (b) and (c) add semantic parsing results either from the input or output side. The upside-down face indicates a negative impact. Our method, \method, introduces semantic hints without perception of the results. 
    }
    \label{fig:sp_method}
    \vspace{-5mm}
\end{figure}

Previous work from \citet{bonial2020dialogue,rapp2022using, Khashabi_Khot_Sabharwal_Roth_2022} demonstrate that integrating semantic parsing results from SRL or AMR parsing into a model's input can effectively enhance its ability to understand illocutionary acts and linguistic abstractions, thereby improving downstream performance.
However, these findings are largely limited to smaller models like BERT \citep{devlin-etal-2019-bert}.
With the rise of Large Language Models (LLMs), it becomes essential to explore how the integration of semantic parsing could impact. Recently, \citet{jin2024analyzing} investigates the role of semantic representation in LLMs by proposing \textsc{AMRCOT}, a method similar to that depicted in Fig.\ref{fig:sp_method} (b). Their findings reveal that introducing AMR results into the input generally harms LLM performance more than it helps, likely because AMR is not yet a representation well-suited for LLMs. However, this analysis remains limited, as it only considers the effects of AMR on several tasks, leaving the broader potential of semantic parsing in LLMs largely unexplored.

In this paper, we systematically investigate the impact of semantic parsing on LLMs to address the question: \textbf{\textit{Can Semantic Information Still Contribute to Improve Downstream Tasks on LLMs?}} We empirically compare different paradigms for integrating semantic parsing into LLMs, as shown in Fig.\ref{fig:sp_method}. These paradigms include approaches commonly used for smaller models, such as incorporating semantic parsing results directly on the input side by fine-tuning or integrating them on the output side. However, these methods negatively affect model performance since they limit fixed types of semantic parsing and might introduce erroneous results.
Thus, we propose a novel prompting approach, \textbf{\method}, illustrated in Fig.\ref{fig:sp_method} (d). Instead of injecting explicit parsing results, \method\ encourages LLMs to harness their internal semantic parsing capabilities through the addition of semantic hints. These hints are as simple as \textbf{``\textit{please use semantic parsing result to enhance comprehension of the sentence's structure and semantics}''}.
Our comprehensive experiments demonstrate that \method promote LLM to focus more on key semantic information, not only achieves superior and consistent performance across various tasks, but also produces more linguistically aligned results, particularly on simplification and paraphrasing tasks, underscoring the effectiveness of semantic parsing for enhancing LLMs' performance.
\section{Semantic Information $\rightarrow$ LLMs}
\begin{figure*}[t]
    \centering
    \includegraphics[width=0.91\textwidth]{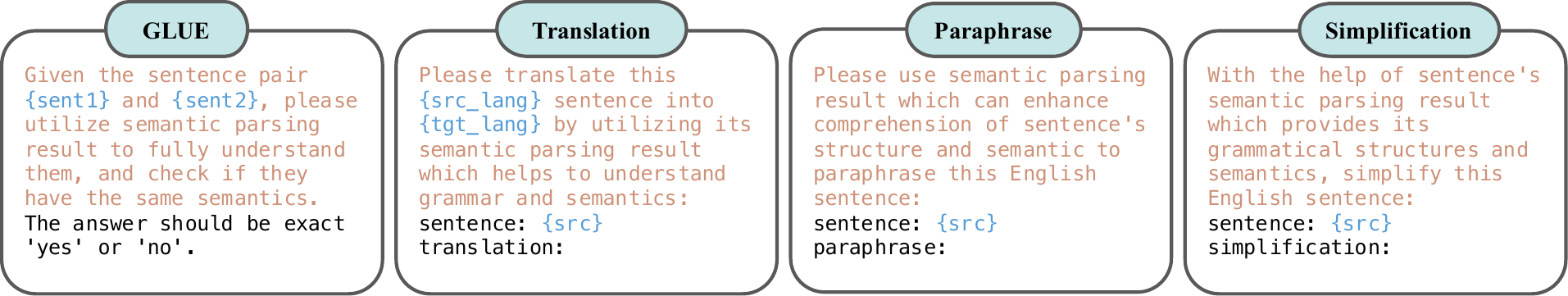}
    \caption{Illustration of \method designed for downstream tasks.
    }
    \label{fig:sense}
    \vspace{-5mm}
\end{figure*}

In this section, we delve into answering the question: \textbf{\textit{Can Semantic Information Still Contribute to Improve Downstream Tasks on LLMs?}}

\subsection{Methodology}
Previous studies, such as those by \citet{ettinger-etal-2023-expert} and \citet{jin2024analyzing}, highlight the difficulty LLMs face in processing the schemes and symbols of explicit semantic parsing results. Their findings suggest that directly integrating these results can degrade model performance. Given that LLMs are already capable of achieving strong results in an end-to-end manner, we propose a novel approach: incorporating semantic parsing hints into the instruction to prompt LLMs to leverage their internal parsing capabilities.

As Fig.\ref{fig:sense} shows, our \method introduces simple semantic hints such as \textit{“utilize semantic parsing result”} to \textit{“fully understand input"} or \textit{"capture grammatical structures and semantics”} to complete downstream tasks. This strategy encourages LLMs to engage in inherent understanding of linguistic structures without requiring explicit semantic parsing results. The workflow outlined in Fig.\ref{fig:sp_method} (d) demonstrates how semantic hints are integrated, and \method works in an zero-shot manner.

\subsection{Datasets and Evaluation}
In our experiments, we select seven understanding tasks from GLUE and three representative generation tasks including Machine Translation, Paraphrasing, and Simplification. 
Specifically, for paraphrasing task, we report three linguistic metrics across lexical, syntactic, and semantic levels, for simplification task, we report SARI and SAMSA which evaluate the predicted simplified sentences from lexical structure and semantic meaning preservation.
More details about our experiments can be found in Appendix \ref{ref:dataset} and \ref{ref:experiment}.

\section{Experimental Results}
\subsection{Main Results}
\paragraph{Results on Understanding Tasks}
From Table \ref{tab:glue}, the results demonstrate that although LLMs currently lag behind smaller models like BERT, the integration of \method significantly narrows this gap. Specifically, \method improves the average performance of GPT-4o-mini from 79.43\% to 81.25\%, bringing it closer to BERT's performance of 83.2\%. Moreover, \method is effective in enhancing the performance of both closed-source models such as GPT-series, and open-source models like LLaMA3. Across all GLUE tasks, \method consistently yields performance gains, with notable improvements in MRPC (72.30\% to 76.47\%), MNLI (73.90\% to 78.20\%) and CoLA (65.49\% to 67.22\%). These results highlight \method's ability to enhance LLMs’ comprehension of input sentences and demonstrate its robustness across diverse tasks.

\begin{table*}[ht]
\centering
\small
\begin{tabular}{lccccccccc}
\toprule
 & SST-2  & MRPC  & QQP   & MNLI  & QNLI  & RTE & CoLA\\
\cmidrule(r){2-8}
\multirow{-2}{*}{System}  & Acc  & Acc  & Acc  & Acc   & Acc  & Acc  & Mcc & \multirow{-2}{*}{Average} \\
\midrule
BERT\textsubscript{LARGE} \citeyearpar{devlin-etal-2019-bert} & 93.20 & 88.00 & 91.30 & 86.60 & 92.30 & 70.40 & 60.60 & 83.20 \\
RoBERTa\textsubscript{LARGE} \citeyearpar{liu2019roberta} & 96.40 & 90.90 & 92.20 & 90.20 & 94.70 & 86.60 & 68.00 & 88.43 \\
\hdashline
LLaMA3-70B & \textbf{95.64}	& 73.52	& 74.60	& 71.90 & 91.30 & 84.48 & 63.90 & 79.34 \\
\rowcolor{blue!5} \textbf{\hspace{3pt} + \method} & 95.18 & 74.04 & 76.50 & 73.10 & 92.80 & 85.56 & 65.53 & 80.25\\
GPT-3.5-turbo & 91.86 & 73.28 & 73.40 & 61.80 & 82.40 & 81.81 & 63.50 & 75.44\\
\rowcolor{blue!5} \textbf{\hspace{3pt} + \method} & 92.20 & 75.49  & \textbf{77.20} & 64.60 & 83.20 & 84.12  & 64.57 & 77.34 \\
GPT-4o-mini & 91.63	& 72.30 & 73.00 & 73.90 & 92.30	& 87.36	& 65.49	& 79.43 \\
\rowcolor{blue!5} \textbf{\hspace{3pt} + \method} & 	92.08 & \textbf{76.47} & 73.00 & \textbf{78.20} & \textbf{93.30} & \textbf{88.45}	& \textbf{67.22} & \textbf{81.25} \\
\bottomrule
\end{tabular}
\caption{Experimental results on GLUE benchmark.}
\label{tab:glue}
\vspace{-5mm}
\end{table*}

\paragraph{Results on Paraphrasing}
\begin{table}[h]
\centering
\small
\begin{tabular}{lccc}
\toprule
  & \multicolumn{3}{c}{Prediction--Source} \\
 \cmidrule(r){2-4}
\multirow{-2}{*}{System}& \makecell{Semantic \\ Similarity ↑} &  \makecell{Lexical \\ Overlap ↓} & \makecell{Syntactic \\ Diversity ↑} \\
\midrule
LLaMA3-70B	& 83.71 & 30.00	& 10.85 \\
\rowcolor{blue!5} \textbf{\hspace{3pt} + \method}	& \textbf{84.02}	& \textbf{29.00}	& \textbf{11.51} \\
GPT-3.5-turbo   & 85.79 & 46.37 & 8.76 \\
\rowcolor{blue!5} \textbf{\hspace{3pt} + \method} & \textbf{85.79} & \textbf{25.33} & \textbf{10.24} \\
GPT-4o-mini	& 89.71	& 39.00	& 7.25 \\
\rowcolor{blue!5} \textbf{\hspace{3pt} + \method}	& \textbf{90.26} & \textbf{34.00}	& \textbf{8.08} \\
\bottomrule
\end{tabular}
\caption{Experimental results on Paraphrasing. We report linguistic metrics between source and prediction.}
\label{tab:para}
\vspace{-6mm}
\end{table}

Table \ref{tab:para} indicates that \method effectively enhances linguistic diversity in paraphrasing tasks while maintaining high semantic similarity. Notably, \method retains the semantic similarity score at 90.26 but significantly reduces lexical overlap from 39.00 to 34.00 and increases syntactic diversity from 7.25 to 8.08. This indicates that the semantic hints introduced by \method lead to more diverse syntactic structures and reduced lexical repetition while preserving the core meaning of the source sentence, which validates the effectiveness of \method in generating paraphrases that are not only semantically faithful but also exhibit greater lexical and syntactic variety.

\paragraph{Results on Simplification}
\begin{table}[h]
\vspace{-4mm}
\centering
\small
\begin{tabular}{lccc}
\toprule
System & BLEU ↑     & SARI ↑     & SAMSA ↑   \\
\midrule
 \multicolumn{4}{c}{TrukCorpus} \\
GPT-3.5-turbo & 58.16   & 42.25    & 31.42  \\
\rowcolor{blue!5} \textbf{\hspace{3pt} + \method}      & \textbf{63.42 }    & \textbf{42.42}     & \textbf{37.03}    \\
\midrule
 \multicolumn{4}{c}{GoogleComp}   \\
\hdashline
GPT-3.5-turbo  & 13.12   & 35.53   & 28.14  \\
\rowcolor{blue!5}  \textbf{\hspace{3pt} + \method} & \textbf{14.31}   & \textbf{35.67}   & \textbf{30.52}  \\
\bottomrule
\end{tabular}
\caption{Experimental results on Simplification. We add two metrics, SARI and SAMSA to evaluate the semantic structure of the output.}
\label{tab:simp}
\vspace{-4mm}
\end{table}

Table \ref{tab:simp} showcases the improved performance of \method on two simplification datasets. Compared to the vanilla prompt, \method delivers higher BLEU scores of 63.42 on TrukCorpus and 14.31 on GoogleComp, alongside a modest increase in SARI, which evaluates the alignment between the source and target sentences. More importantly, the SAMSA scores, which measure the preservation of syntactic structure, show substantial improvement, reaching 37.03 and 30.52 respectively. These results demonstrate that integrating semantic hints into prompts enhances the model's ability to simplify sentences while preserving their original structure, resulting in more effective overall simplification.

\vspace{-2mm}
\paragraph{Results on Machine Translation}

We further conduct experiments on Machine Translation task and present a comparative analysis of GPT-3.5-turbo across the vanilla prompt, our \method, and other state-of-the-art systems in Table \ref{tab:mt}.  Results show that \method consistently enhances GPT-3.5 across all evaluated metrics and language pairs. For the DE-EN task, \method achieves the highest scores: COMET22 (86.44), ChrF (59.08), and BLEU (33.75), outperforming the WMT-Best system. Similarly, in the EN-DE task, \method significantly boosts GPT-3.5’s performance, reaching COMET22 (86.65), ChrF (62.84), and BLEU (34.18). These improvements highlight the effectiveness of \method in enhancing GPT-3.5's ability to handle translation tasks across different language pairs. The results for ZH-EN and EN-ZH in Table \ref{tab:mt} further confirm \method's effectiveness.

\subsection{Analytical Results}

\paragraph{Analysis of Different Paradigms}
In Table \ref{analy:parsing}, we compare various approaches for incorporating semantic parsing into LLMs. We examine methods that either concatenate pre-generated parsing results using LLM or generate them on output side\footnote{We do not specify certain type of semantic parsing during our experiments.}. The results demonstrate that directly adding semantic parsing results degrades performance, aligning with findings by \citet{jin2024analyzing}. This degradation arises from the unfamiliar symbolic representation and the diversity of semantic parsing tasks, integrating specific type, and potentially erroneous results limits LLM’s capability. In contrast, \method avoids explicit incorporation while consistently outperforming these methods. Such finding underscores \method as a more effective strategy for leveraging semantic parsing on LLMs.

\begin{table*}[t]
\centering
\small
   \resizebox{0.98\linewidth}{!}{
    \begin{tabular}{lccccccc}
    \toprule
    System & SST-2  & MRPC  & QQP   & MNLI  & QNLI  & RTE & CoLA \\
    \midrule
    GPT-3.5-turbo & 91.86 & 73.28 & 73.40 & 61.80 & 82.40 & 81.81 & 63.50  \\
    \hspace{3pt} + CoT \citeyearpar{NEURIPS2022_8bb0d291} & 89.11$_{\textcolor{green}{-2.75}}$ & 73.28$_{\textcolor{red}{+0.00}}$ & 77.00$_{\textcolor{red}{+3.60}}$ & 56.20$_{\textcolor{green}{-5.60}}$ & 82.70$_{\textcolor{red}{+0.30}}$ & 82.54$_{\textcolor{red}{+0.73}}$ & 64.32$_{\textcolor{red}{+0.82}}$ \\
    \hspace{3pt} + SP-Input & 87.50$_{\textcolor{green}{-4.36}}$ & 74.26$_{\textcolor{red}{+0.98}}$ & 74.30$_{\textcolor{red}{+0.90}}$ & 50.50$_{\textcolor{green}{-11.30}}$ & 78.40$_{\textcolor{green}{-4.00}}$ & 84.11$_{\textcolor{red}{+2.30}}$ & 58.37$_{\textcolor{green}{-5.13}}$ \\
    \hspace{3pt} + SP-Output & 89.11$_{\textcolor{green}{-2.75}}$ & 73.52$_{\textcolor{red}{+0.24}}$ & 71.90$_{\textcolor{green}{-1.50}}$ & 62.00$_{\textcolor{red}{+0.20}}$ & 78.40$_{\textcolor{green}{-4.00}}$ & 81.59$_{\textcolor{green}{-0.22}}$ & 64.44$_{\textcolor{red}{+0.94}}$  \\
    \rowcolor{blue!5} \textbf{\hspace{3pt} + \method} & 92.20$_{\textcolor{red}{+0.34}}$ & 75.49$_{\textcolor{red}{+2.21}}$ & 77.20$_{\textcolor{red}{+3.80}}$ & 64.60$_{\textcolor{red}{+2.80}}$ & 83.20$_{\textcolor{red}{+0.80}}$ & 84.12$_{\textcolor{red}{+2.31}}$ & 64.57$_{\textcolor{red}{+1.07}}$ \\
    \bottomrule
    \end{tabular}
    }
    \caption{Analysis of different approaches that introduce semantic parsing into LLMs on GLUE benchmark. Improvements are marked in \textcolor{red}{red} and decreases in \textcolor{green}{green}, relative to GPT-3.5-turbo.}
    \vspace{-3mm}
\label{analy:parsing}
\end{table*}

\begin{figure*}[t]
    \centering
    \includegraphics[width=0.75\textwidth]{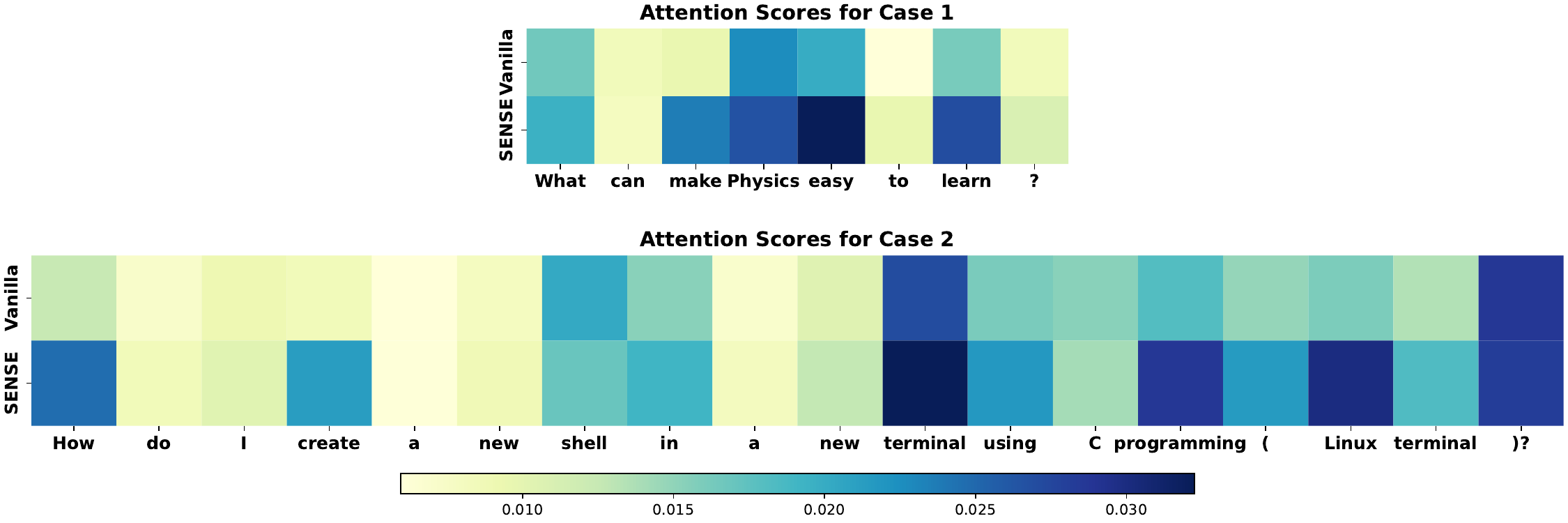}
    \caption{Visualization of attention scores from LLaMA3-70B on the source sentence in the Paraphrasing Task.
    }
    \label{fig:visual}
    \vspace{-3mm}
\end{figure*}

\vspace{-0.5mm}
\paragraph{Comparison with Chain-of-Thought}
Since \method shares similarities with CoT \citep{NEURIPS2022_8bb0d291}, which works by adding "Let's think step by step", we compare it on GLUE (Table \ref{analy:parsing}) and machine translation task (Table \ref{tab:mt}). While CoT degrades performance across tasks, as it is better suited for reasoning tasks, \method significantly enhances LLM performance by improving the model's ability to understand input sentences, thus yielding better results.

\vspace{-0.5mm}
\paragraph{Visualization of Attention Scores}
We present the distribution of attention scores for paraphrasing task in Fig.\ref{fig:visual}, where we average attention scores for each output token with respect to original sentence. The visualization reveals that, compared to vanilla prompt, \method places greater emphasis on key semantic elements, such as important lexical units and core components. This indicates that \method more effectively directs attention toward critical semantic information, and thus generates outputs that are more linguistic-aligned. 
Additionally, we provide case study on such examples in Table \ref{tab:case} and \ref{tab:case_nli}. While both vanilla prompt and \method successfully capture the paraphrased meaning, \method is superior at transforming syntactical structure and utilizing more diverse expressions.
\section{Related Work}
Semantic parsing has significantly contributed to enhancing the performance of smaller language models. Integrating results from SRL and AMR \cite{gildea2002automatic, Palmer_Titov_Wu_2010, Banarescu_Bonial_Cai_Georgescu_Griffitt_Hermjakob_Knight_Koehn_Palmer_Schneider_2013} has shown to improve model performance on various tasks \cite{Khashabi_Khot_Sabharwal_Roth_2022, rapp2022using, xu2020semantic, si-etal-2022-mining, si2024spokenwoz}.
However, the effectiveness of semantic parsing to LLMs is under-explored. Recent work, such as \citet{jin2024analyzing}, explores the use of AMR results with LLMs and finds that direct integration of these results may not always yield positive influences. 
Unlike approaches focused on optimizing prompts directly \cite{zhou2022large, pryzant2023automatic, deng2022rlprompt, guo2024connecting}, our work proposes a novel strategy for leveraging semantic parsing in LLMs. Similar to CoT \cite{wei2022chain,NEURIPS2022_8bb0d291} and DTG \cite{li2023deliberate}, our method involves integrating semantic parsing hints into prompts rather than optimizing the prompts. 

\section{Conclusion}
In this paper, we rethink leveraging semantic parsing to enhance LLMs' performance. Contrary to smaller models, where direct integration of parsing results can be beneficial, we find that this negatively impacts LLMs. With the help of our proposed \method, which introduces semantic hints within prompts, LLMs can better comprehend input sentences. Experiments show that \method\ achieves great performance across both understanding and generation tasks, and helps models capture lexical and syntactic structures, producing outputs that align more closely with linguistic metrics.

\section*{Limitations}
While we validate the effectiveness of \method\ across both understanding and generation tasks, there are limitations that remain for future exploration:
Firstly, our validation is restricted to the LLaMA and GPT-series models. Extending \method\ to other LLM architectures will be necessary to confirm its general applicability.
Secondly, although \method\ shows promising results on a range of NLP tasks, its performance across more diverse datasets and applications needs further investigation. Our experiments focus on tasks where the benefits of semantic parsing have been established, but broader testing is required to fully assess its potential.
Additionally, the underlying mechanism of how semantic parsing influences LLM decision-making remains unclear, as LLMs function largely as black-box systems.  Our validation primarily involves comparing methods that directly incorporate semantic parsing results from the input or output sides, and analyzing the outputs in contrast to both the vanilla prompt and \method.

\section*{Acknowledgements}
We thank all reviewers for their great efforts. This work is supported by the National Science Foundation of China under Grant No.61876004 and 61936012.

\bibliography{custom}

\begin{thebibliography}{29}
\providecommand{\natexlab}[1]{#1}

\bibitem[{An et~al.(2023)An, Zheng, Gao, Zhao, and Chang}]{an-etal-2023-coarse}
Kaikai An, Ce~Zheng, Bofei Gao, Haozhe Zhao, and Baobao Chang. 2023.
\newblock \href {https://doi.org/10.18653/v1/2023.findings-emnlp.897} {Coarse-to-fine dual encoders are better frame identification learners}.
\newblock In \emph{Findings of the Association for Computational Linguistics: EMNLP 2023}, pages 13455--13466, Singapore. Association for Computational Linguistics.

\bibitem[{Baker et~al.(2007)Baker, Ellsworth, and Erk}]{baker-etal-2007-semeval}
Collin Baker, Michael Ellsworth, and Katrin Erk. 2007.
\newblock \href {https://aclanthology.org/S07-1018} {{S}em{E}val-2007 task 19: Frame semantic structure extraction}.
\newblock In \emph{Proceedings of the Fourth International Workshop on Semantic Evaluations ({S}em{E}val-2007)}, pages 99--104, Prague, Czech Republic. Association for Computational Linguistics.

\bibitem[{Banarescu et~al.(2013)Banarescu, Bonial, Cai, Georgescu, Griffitt, Hermjakob, Knight, Koehn, Palmer, and Schneider}]{Banarescu_Bonial_Cai_Georgescu_Griffitt_Hermjakob_Knight_Koehn_Palmer_Schneider_2013}
Laura Banarescu, Claire Bonial, Shu Cai, Madalina Georgescu, Kira Griffitt, Ulf Hermjakob, Kevin Knight, Philipp Koehn, Martha Palmer, and Nathan Schneider. 2013.
\newblock Abstract meaning representation for sembanking.
\newblock \emph{Linguistic Annotation Workshop,Linguistic Annotation Workshop}.

\bibitem[{Bonial et~al.(2020)Bonial, Donatelli, Abrams, Lukin, Tratz, Marge, Artstein, Traum, and Voss}]{bonial2020dialogue}
Claire Bonial, Lucia Donatelli, Mitchell Abrams, Stephanie Lukin, Stephen Tratz, Matthew Marge, Ron Artstein, David Traum, and Clare Voss. 2020.
\newblock Dialogue-amr: abstract meaning representation for dialogue.
\newblock In \emph{Proceedings of the Twelfth Language Resources and Evaluation Conference}, pages 684--695.

\bibitem[{Deng et~al.(2022)Deng, Wang, Hsieh, Wang, Guo, Shu, Song, Xing, and Hu}]{deng2022rlprompt}
Mingkai Deng, Jianyu Wang, Cheng-Ping Hsieh, Yihan Wang, Han Guo, Tianmin Shu, Meng Song, Eric~P Xing, and Zhiting Hu. 2022.
\newblock Rlprompt: Optimizing discrete text prompts with reinforcement learning.
\newblock \emph{arXiv preprint arXiv:2205.12548}.

\bibitem[{Devlin et~al.(2019)Devlin, Chang, Lee, and Toutanova}]{devlin-etal-2019-bert}
Jacob Devlin, Ming-Wei Chang, Kenton Lee, and Kristina Toutanova. 2019.
\newblock \href {https://doi.org/10.18653/v1/N19-1423} {{BERT}: Pre-training of deep bidirectional transformers for language understanding}.
\newblock In \emph{Proceedings of the 2019 Conference of the North {A}merican Chapter of the Association for Computational Linguistics: Human Language Technologies, Volume 1 (Long and Short Papers)}, pages 4171--4186, Minneapolis, Minnesota. Association for Computational Linguistics.

\bibitem[{Ettinger et~al.(2023)Ettinger, Hwang, Pyatkin, Bhagavatula, and Choi}]{ettinger-etal-2023-expert}
Allyson Ettinger, Jena Hwang, Valentina Pyatkin, Chandra Bhagavatula, and Yejin Choi. 2023.
\newblock \href {https://doi.org/10.18653/v1/2023.findings-emnlp.553} {{``}you are an expert linguistic annotator{''}: Limits of {LLM}s as analyzers of {A}bstract {M}eaning {R}epresentation}.
\newblock In \emph{Findings of the Association for Computational Linguistics: EMNLP 2023}, pages 8250--8263, Singapore. Association for Computational Linguistics.

\bibitem[{Gildea and Jurafsky(2002)}]{gildea2002automatic}
Daniel Gildea and Dan Jurafsky. 2002.
\newblock Automatic labeling of semantic roles.
\newblock \emph{Computational Linguistics}, 28(3):245--288.

\bibitem[{Guo et~al.(2024)Guo, Wang, Guo, Li, Song, Tan, Liu, Bian, and Yang}]{guo2024connecting}
Qingyan Guo, Rui Wang, Junliang Guo, Bei Li, Kaitao Song, Xu~Tan, Guoqing Liu, Jiang Bian, and Yujiu Yang. 2024.
\newblock \href {https://arxiv.org/abs/2309.08532} {Connecting large language models with evolutionary algorithms yields powerful prompt optimizers}.
\newblock \emph{Preprint}, arXiv:2309.08532.

\bibitem[{Hendy et~al.(2023)Hendy, Abdelrehim, Sharaf, Raunak, Gabr, Matsushita, Kim, Afify, and Awadalla}]{hendy2023good}
Amr Hendy, Mohamed Abdelrehim, Amr Sharaf, Vikas Raunak, Mohamed Gabr, Hitokazu Matsushita, Young~Jin Kim, Mohamed Afify, and Hany~Hassan Awadalla. 2023.
\newblock How good are gpt models at machine translation? a comprehensive evaluation.
\newblock \emph{arXiv preprint arXiv:2302.09210}.

\bibitem[{Huang et~al.(2023)Huang, Iyer, Hsu, Kumar, Chang, and Galstyan}]{huang-etal-2023-paraamr}
Kuan-Hao Huang, Varun Iyer, I-Hung Hsu, Anoop Kumar, Kai-Wei Chang, and Aram Galstyan. 2023.
\newblock \href {https://doi.org/10.18653/v1/2023.acl-long.447} {{P}ara{AMR}: A large-scale syntactically diverse paraphrase dataset by {AMR} back-translation}.
\newblock In \emph{Proceedings of the 61st Annual Meeting of the Association for Computational Linguistics (Volume 1: Long Papers)}, pages 8047--8061, Toronto, Canada. Association for Computational Linguistics.

\bibitem[{Jin et~al.(2024)Jin, Chen, Gonzalez, Liu, Zhang, Michael, Sch{\"o}lkopf, and Diab}]{jin2024analyzing}
Zhijing Jin, Yuen Chen, Fernando Gonzalez, Jiarui Liu, Jiayi Zhang, Julian Michael, Bernhard Sch{\"o}lkopf, and Mona Diab. 2024.
\newblock Analyzing the role of semantic representations in the era of large language models.
\newblock \emph{arXiv preprint arXiv:2405.01502}.

\bibitem[{Khashabi et~al.(2022)Khashabi, Khot, Sabharwal, and Roth}]{Khashabi_Khot_Sabharwal_Roth_2022}
Daniel Khashabi, Tushar Khot, Ashish Sabharwal, and Dan Roth. 2022.
\newblock \href {https://doi.org/10.1609/aaai.v32i1.11574} {Question answering as global reasoning over semantic abstractions}.
\newblock \emph{Proceedings of the AAAI Conference on Artificial Intelligence}, 32(1).

\bibitem[{Kojima et~al.(2022)Kojima, Gu, Reid, Matsuo, and Iwasawa}]{NEURIPS2022_8bb0d291}
Takeshi Kojima, Shixiang~(Shane) Gu, Machel Reid, Yutaka Matsuo, and Yusuke Iwasawa. 2022.
\newblock \href {https://proceedings.neurips.cc/paper_files/paper/2022/file/8bb0d291acd4acf06ef112099c16f326-Paper-Conference.pdf} {Large language models are zero-shot reasoners}.
\newblock In \emph{Advances in Neural Information Processing Systems}, volume~35, pages 22199--22213. Curran Associates, Inc.

\bibitem[{Li et~al.(2023)Li, Wang, Guo, Song, Tan, Hassan, Menezes, Xiao, Bian, and Zhu}]{li2023deliberate}
Bei Li, Rui Wang, Junliang Guo, Kaitao Song, Xu~Tan, Hany Hassan, Arul Menezes, Tong Xiao, Jiang Bian, and JingBo Zhu. 2023.
\newblock Deliberate then generate: Enhanced prompting framework for text generation.
\newblock \emph{arXiv preprint arXiv:2305.19835}.

\bibitem[{Liu et~al.(2019)Liu, Ott, Goyal, Du, Joshi, Chen, Levy, Lewis, Zettlemoyer, and Stoyanov}]{liu2019roberta}
Yinhan Liu, Myle Ott, Naman Goyal, Jingfei Du, Mandar Joshi, Danqi Chen, Omer Levy, Mike Lewis, Luke Zettlemoyer, and Veselin Stoyanov. 2019.
\newblock Roberta: A robustly optimized bert pretraining approach.
\newblock \emph{arXiv preprint arXiv:1907.11692}.

\bibitem[{{OpenAI}(2023)}]{OpenAI2023ChatGPT}
{OpenAI}. 2023.
\newblock Chatgpt: Optimizing language models for dialogue.
\newblock \url{https://openai.com/blog/chatgpt}.
\newblock Accessed: 2023-04-01.

\bibitem[{Palmer et~al.(2010)Palmer, Titov, and Wu}]{Palmer_Titov_Wu_2010}
Martha Palmer, Ivan Titov, and Shumin Wu. 2010.
\newblock Semantic role labeling.
\newblock \emph{Computational Linguistics}.

\bibitem[{Pryzant et~al.(2023)Pryzant, Iter, Li, Lee, Zhu, and Zeng}]{pryzant2023automatic}
Reid Pryzant, Dan Iter, Jerry Li, Yin~Tat Lee, Chenguang Zhu, and Michael Zeng. 2023.
\newblock Automatic prompt optimization with" gradient descent" and beam search.
\newblock \emph{arXiv preprint arXiv:2305.03495}.

\bibitem[{Rapp(2022)}]{rapp2022using}
Reinhard Rapp. 2022.
\newblock Using semantic role labeling to improve neural machine translation.
\newblock In \emph{Proceedings of the Thirteenth Language Resources and Evaluation Conference}, pages 3079--3083.

\bibitem[{Rei et~al.(2022)Rei, De~Souza, Alves, Zerva, Farinha, Glushkova, Lavie, Coheur, and Martins}]{rei2022comet}
Ricardo Rei, Jos{\'e}~GC De~Souza, Duarte Alves, Chrysoula Zerva, Ana~C Farinha, Taisiya Glushkova, Alon Lavie, Luisa Coheur, and Andr{\'e}~FT Martins. 2022.
\newblock Comet-22: Unbabel-ist 2022 submission for the metrics shared task.
\newblock In \emph{Proceedings of the Seventh Conference on Machine Translation (WMT)}, pages 578--585.

\bibitem[{Si et~al.(2024)Si, Ma, Gao, Wu, Lin, Dai, Li, Yan, Huang, and Li}]{si2024spokenwoz}
Shuzheng Si, Wentao Ma, Haoyu Gao, Yuchuan Wu, Ting-En Lin, Yinpei Dai, Hangyu Li, Rui Yan, Fei Huang, and Yongbin Li. 2024.
\newblock Spokenwoz: A large-scale speech-text benchmark for spoken task-oriented dialogue agents.
\newblock \emph{Advances in Neural Information Processing Systems}, 36.

\bibitem[{Si et~al.(2022)Si, Zeng, and Chang}]{si-etal-2022-mining}
Shuzheng Si, Shuang Zeng, and Baobao Chang. 2022.
\newblock \href {https://doi.org/10.18653/v1/2022.naacl-main.356} {Mining clues from incomplete utterance: A query-enhanced network for incomplete utterance rewriting}.
\newblock In \emph{Proceedings of the 2022 Conference of the North American Chapter of the Association for Computational Linguistics: Human Language Technologies}, pages 4839--4847, Seattle, United States. Association for Computational Linguistics.

\bibitem[{Sulem et~al.(2018)Sulem, Abend, and Rappoport}]{sulem2018semantic}
Elior Sulem, Omri Abend, and Ari Rappoport. 2018.
\newblock Semantic structural evaluation for text simplification.
\newblock \emph{arXiv preprint arXiv:1810.05022}.

\bibitem[{Wang et~al.(2019)Wang, Singh, Michael, Hill, Levy, and Bowman}]{wang2019gluemultitaskbenchmarkanalysis}
Alex Wang, Amanpreet Singh, Julian Michael, Felix Hill, Omer Levy, and Samuel~R. Bowman. 2019.
\newblock \href {https://arxiv.org/abs/1804.07461} {Glue: A multi-task benchmark and analysis platform for natural language understanding}.
\newblock \emph{Preprint}, arXiv:1804.07461.

\bibitem[{Wei et~al.(2022)Wei, Wang, Schuurmans, Bosma, Xia, Chi, Le, Zhou et~al.}]{wei2022chain}
Jason Wei, Xuezhi Wang, Dale Schuurmans, Maarten Bosma, Fei Xia, Ed~Chi, Quoc~V Le, Denny Zhou, et~al. 2022.
\newblock Chain-of-thought prompting elicits reasoning in large language models.
\newblock \emph{Advances in neural information processing systems}, 35:24824--24837.

\bibitem[{Xu et~al.(2020)Xu, Tan, Song, Wu, Zhang, Song, and Yu}]{xu2020semantic}
Kun Xu, Haochen Tan, Linfeng Song, Han Wu, Haisong Zhang, Linqi Song, and Dong Yu. 2020.
\newblock Semantic role labeling guided multi-turn dialogue rewriter.
\newblock In \emph{Proceedings of the 2020 Conference on Empirical Methods in Natural Language Processing (EMNLP)}, pages 6632--6639.

\bibitem[{Zhang et~al.(2023)Zhang, Fang, Zhang, Ma, Zhou, Huang, Bu, Gui, Chen, Chen et~al.}]{zhang2023bayling}
Shaolei Zhang, Qingkai Fang, Zhuocheng Zhang, Zhengrui Ma, Yan Zhou, Langlin Huang, Mengyu Bu, Shangtong Gui, Yunji Chen, Xilin Chen, et~al. 2023.
\newblock Bayling: Bridging cross-lingual alignment and instruction following through interactive translation for large language models.
\newblock \emph{arXiv preprint arXiv:2306.10968}.

\bibitem[{Zhou et~al.(2022)Zhou, Muresanu, Han, Paster, Pitis, Chan, and Ba}]{zhou2022large}
Yongchao Zhou, Andrei~Ioan Muresanu, Ziwen Han, Keiran Paster, Silviu Pitis, Harris Chan, and Jimmy Ba. 2022.
\newblock Large language models are human-level prompt engineers.
\newblock \emph{arXiv preprint arXiv:2211.01910}.

\end{thebibliography}

\newpage
\appendix
\section{Supplementary Details}
\subsection{Details about Datasets}
\label{ref:dataset}
We list the details of each dataset, including source, number, and metrics for each task in Table \ref{tab:dataset}, and we sample a subset of data if the original dataset is large to reduce the API cost.

\begin{table}[h]
    \centering
    \small
    \begin{tabular}{lccc}
    \toprule
    Dataset & Num. & Metrics \\
    \midrule
    SST-2  & 872 & Acc \\
    MRPC  & 408 & Acc \\
    QQP  & 1000 & Acc \\
    MNLI  & 1000 & Acc \\
    QNLI  & 1000 & Acc \\
    RTE  & 277 & Acc \\
    CoLA  & 1053 & Mcc \\
    WMT DE-EN & 1984 &  BLEU, COMET22, Chrf\\
    WMT EN-DE & 1875 &  BLEU, COMET22, Chrf\\
    WMT ZH-EN & 1875 &  BLEU, COMET22, Chrf\\
    WMT EN-ZH & 1875 &  BLEU, COMET22, Chrf\\ 
    QQP & 2500 & Lexical, Syntactic, Semantic\\
    TurkCorpus & 359 & BLEU, SARI, SAMSA \\
    GoogleComp & 1000 & BLEU, SARI, SAMSA \\
    \bottomrule
    \end{tabular}
    \caption{Statistics of the dataset we use in our experiment.}
    \label{tab:dataset}
    \vspace{-5mm}
\end{table}

\paragraph{GLUE}
We test on seven tasks from GLUE benchmark \cite{wang2019gluemultitaskbenchmarkanalysis} and report the Matthews Correlation Coefficient (MCC) for CoLA and Accuracy (Acc) for the left tasks.

\paragraph{Machine Translation}
For machine translation, we evaluate our method on the WMT22 \footnote{\url{https://machinetranslate.org/wmt22}} dataset, focusing on two language pairs: EN-DE (English to German)
EN-ZH (English to Chinese) and report COMET22 \citep{rei2022comet}, CHRF, and BLEU scores \footnote{BLEU+case.mixed+numrefs.1+smooth.exp+tok.13a}.

\paragraph{Paraphrasing}
We evaluate on Quora Question Pairs (QQP) \footnote{\url{https://quoradata.quora.com/First-Quora-Dataset-Release-Question-Pairs}} dataset. To analyze results professionally, we follow \citet{huang-etal-2023-paraamr} and report three linguistic evaluation metrics across lexical, syntactic, and semantic levels.

\paragraph{Simplification}
For text simplification, we evaluate on TurkCorpus and GoogleComp and use BLEU, SARI, and SAMSA as the evaluation metrics. 
Specifically, SARI \footnote{\url{https://huggingface.co/spaces/evaluate-metric/sari}} (System output Against References and against the Input sentence) is used to compare the predicted simplified sentences against the reference and the source sentences, and SAMSA \citep{sulem2018semantic} is a metric specifically designed for text simplification that evaluates structural simplification and meaning preservation.

\subsection{Details about Experiment}
\label{ref:experiment}
\subsubsection{Experimental Setup}
We test our \method on GPT-3.5-turbo, GPT-4o-mini \citep{OpenAI2023ChatGPT} with the version of 2023-11-06 and 2024-07-18, and LLaMA3-70B-Instruct \footnote{\url{https://llama.meta.com/docs/model-cards-and-prompt-formats/meta-llama-3}}. The temperature is set to 0 and top\_p set to 1.

\subsubsection{Prompts used in Experiments}
We release the prompts we use during our experiments in Table \ref{tab:prompt1} and Table \ref{tab:prompt2}.

\begin{table*}
    \small
    \begin{tabular}{p{0.1\linewidth}p{0.1\linewidth}p{0.8\linewidth}}
    \toprule
    Dataset & Method & Prompt \\
    \midrule
    \multirow{3}{*}{SST-2} & Vanilla & Given this sentence: \{sentence\}, please classify its sentiment as positive or negative. The answer should be exactly 'positive' or 'negative'. \\
    & CoT & Given this sentence: \{sentence\}, please think step by step, and then classify its sentiment as positive or negative. The answer should be exactly 'positive' or 'negative'. \\
    & SP-Input & Given this sentence: \{sentence\} and its semantic parsing result \{parsing\}, please classify the sentence's sentiment as positive or negative. The answer should be exactly 'positive' or 'negative'.  \\
    & SP-Output & Given this sentence: \{sentence\}, please first parse this sentence and then classify the sentence's sentiment as positive or negative. The answer should be exactly 'positive' or 'negative'. \\
    & \textbf{\method} & Given this sentence: \{sentence\}, please use semantic parsing result which can enhance comprehension of the sentence's structure and semantics to classify the sentence's sentiment. The answer should be exactly 'positive' or 'negative'. \\
    \midrule
    \multirow{3}{*}{MRPC} & Vanilla & Given the sentence pair \{sentence1\} and \{sentence2\}, please check if these two sentences have the same semantics. The answer should be exactly 'yes' or 'no'. \\
    & CoT & Given the sentence pair \{sentence1\} and \{sentence2\}, please think step by step, and then check if these two sentences have the same semantics. The answer should be exactly 'yes' or 'no'. \\
    & SP-Input & Given the sentence pair \{sentence1\} and \{sentence2\} and their semantic parsing results \{parsing1\} and \{parsing2\}, please check if these two sentences have the same semantics. The answer should be exactly 'yes' or 'no'. \\
    & SP-Output & Given the sentence pair \{sentence1\} and \{sentence2\}, please first parse these sentences and then check if these two sentences have the same semantics. The answer should be exactly 'yes' or 'no'.  \\
    & \textbf{\method} & Given the sentence pair \{sentence1\} and \{sentence2\}, please use semantic parsing result which can enhance comprehension of the sentence's structure and semantics to measure if these two sentences have the same semantics. The answer should be exactly 'yes' or 'no'.  \\
    \midrule
    \multirow{3}{*}{MNLI} & Vanilla & Given the sentence1 \{premise\} and sentence2 \{hypothesis\}, determine whether sentence2 entail, contradict, or is it neutral to sentence1. The answer should be exactly 'entail' or 'contradict' or 'neutral'.\\
    & CoT & Given the sentence1 \{premise\} and sentence2 \{hypothesis\}, please think step by step, and then determine whether sentence2 entail, contradict, or is it neutral to sentence1. The answer should be exactly 'entail' or 'contradict' or 'neutral'. \\
    & SP-Input & Given the sentence1 \{premise\} and sentence2 \{hypothesis\} and their semantic parsing results \{parsing1\} and \{parsing2\}, please determine whether sentence2 entail, contradict, or is it neutral to sentence1. The answer should be exactly 'entail' or 'contradict' or 'neutral'. \\
    & SP-Output & Given the sentence1 \{premise\} and sentence2 \{hypothesis\}, please first parse these sentence to fully understand its structure and semantics and then determine whether sentence1 entail, contradict, or is neutral to sentence2. The answer should be exactly 'entail' or 'contradict' or 'neutral'. \\
    & \textbf{\method} & Given the sentence1 \{premise\} and sentence2 \{hypothesis\}, please use semantic parsing result which can enhance comprehension of the sentence's structure and semantics to determine whether sentence1 entail, contradict, or is neutral to sentence2. The answer should be exactly 'entail' or 'contradict' or 'neutral'. \\
    \midrule
    \multirow{3}{*}{QNLI} & Vanilla & Given the sentence1 \{question\} and sentence2 \{sentence\}, please determine if the sentence contains the answer to the question. The answer should be exactly 'entail' or 'not entail'.\\
    & CoT & Given the sentence1 \{question\} and sentence2 \{sentence\}, please think step by step, and then determine if the sentence contains the answer to the question. The answer should be exactly 'entail' or 'not entail'. \\
    & SP-Input & Given the sentence1 \{question\} and sentence2 \{sentence\} and their semantic parsing results \{parsing1\} and \{parsing2\}, please determine if the sentence contains the answer to the question. The answer should be exactly 'entail' or 'not entail'. \\
    & SP-Output & Given the sentence1 \{question\} and sentence2 \{sentence\}, please first parse these sentences and then determine if the sentence contains the answer to the question. The answer should be exactly 'entail' or 'not entail'. \\
    & \textbf{\method} & Given the sentence1 \{question\} and sentence2 \{sentence\}, please use semantic parsing result which can enhance comprehension of the sentence's structure and semantics to determine if the sentence contains the answer to the question. The answer should be exactly 'entail' or 'not entail'. \\
    \midrule
    \multirow{3}{*}{CoLA} & Vanilla & Given the sentence: \{sentence\}, please check if the sentence is grammatically correct. The answer should be exactly 'yes' or 'no'. \\
    & CoT & Given the sentence: \{sentence\}, please think step by step, and then check if the sentence is grammatically correct. The answer should be exactly 'yes' or 'no'.\\
    & SP-Input & Given the sentence: \{sentence\} and its semantic parsing result \{parsing\}, please check if the sentence is grammatically correct. The answer should be exactly 'yes' or 'no'. \\
    & SP-Output & Given the sentence: \{sentence\}, please first parse this sentence and then check if the sentence is grammatically correct. The answer should be exactly 'yes' or 'no'. \\
    & \textbf{\method} & Given the sentence: \{sentence\}, please use semantic parsing result which can enhance comprehension of the sentence's structure and semantics to check if the sentence is grammatically correct. The answer should be exactly 'yes' or 'no'. \\
    \bottomrule
    \end{tabular}
    \caption{We list the prompts we use during our experiments on GLUE benchmarks and omit QQP and RTE since QQP is similar to MRPC and RTE is similar to MNLI.}
    \label{tab:prompt1}
\end{table*}

\begin{table*}
    \small
    \begin{tabular}{p{0.1\linewidth}p{0.1\linewidth}p{0.8\linewidth}}
    \toprule
    Dataset & Method & Prompt \\
    \midrule
    \multirow{2}{*}{WMT22} & Vanilla & Please translate this \{src\_lang\} sentence into \{tgt\_lang\}: sentence: \{src\} translation: \\
    & \textbf{\method} & Please translate this \{src\_lang\} sentence into \{tgt\_lang\} by utilizing its semantic parsing result which helps to understand grammar and semantics: sentence: \{src\} translation: \\
    \midrule
    \multirow{2}{*}{Simplification} & Vanilla & Please simplify this English sentence: sentence: \{src\} simplification: \\
    & \textbf{\method} & With the help of the sentence's semantic parsing result which provides its grammatical structures and semantics, simplify this English sentence: sentence: \{src\} simplification:  \\
    \midrule
    \multirow{2}{*}{Paraphrasing} & Vanilla & Please paraphrase this English sentence: sentence: \{src\} paraphrase:\\
    & \textbf{\method} & Please use semantic parsing result which can enhance comprehension of sentence's structure and semantic to paraphrase this English sentence: sentence: \{src\} paraphrase: \\
    \bottomrule
    \end{tabular}
    \caption{We list the prompts we use during our experiments on generation tasks.}
    \label{tab:prompt2}
\end{table*}

\subsection{Additional Experimental Results}
\paragraph{Results on WMT22}
From Table \ref{tab:mt}, for the ZH-EN translation task, \method improves GPT-3.5-turbo's ChrF (58.50) and BLEU (27.04) scores, though the COMET22 score (80.47) is slightly lower than the baseline. In the EN-ZH task, \method achieves the highest COMET22 (88.06) and enhances ChrF (39.86) and BLEU (44.40) compared to baselines. 

\begin{table*}[h]
\centering
\small
\begin{tabular}{lcccccc}
    \toprule
    & \multicolumn{3}{c}{DE-EN} & \multicolumn{3}{c}{EN-DE} \\
    \cmidrule(r){2-4} \cmidrule(r){5-7}
    \multirow{-2}{*}{System} & COMET22 ↑   & ChrF ↑    & BLEU ↑    & COMET22 ↑   & Chrf ↑     & BLEU ↑      \\
    \midrule
    WMT-Best & 85.00 & 58.50 & 33.40 & 87.20 & 64.60 & 38.40 \\
    \hdashline
    GPT EVAL \citeyearpar{hendy2023good} & 84.80 & 58.30 & 33.40 &  84.20 & 59.60 & 30.90 \\
    DTG 5-shot \citeyearpar{li2023deliberate}  & 85.40 & 58.20 &  33.20  & 86.30 & 61.60 & 33.40 \\
    BayLing \citeyearpar{zhang2023bayling}  & 85.47 & 58.65 & 32.94   & \textbf{86.93} & 62.76 & 34.12 \\
    \hdashline
    GPT-3.5-turbo & 85.71 & 58.19 & 33.15  & 84.60 & 60.48 & 33.42 \\
    + CoT &	84.99 & 57.74 & 31.46 & 84.95 & 61.17 & 29.70   \\
    \rowcolor{blue!5} \textbf{\hspace{3pt} + \method} & \textbf{86.44} & \textbf{59.08} & \textbf{33.75}  & 86.65  & \textbf{62.84} & \textbf{34.18} \\
    \midrule
    & \multicolumn{3}{c}{ZH-EN} & \multicolumn{3}{c}{EN-ZH} \\
    \cmidrule(r){2-4} \cmidrule(r){5-7}
    \multirow{-2}{*}{System} & COMET22 ↑   & ChrF ↑    & BLEU ↑    & COMET22 ↑   & Chrf ↑     & BLEU ↑      \\
    \midrule
    WMTBest & 81.00 & 61.10 & 33.50 & 86.70 &  41.10 &  44.80 \\
    \hdashline
    GPT EVAL \citeyearpar{hendy2023good}  & 81.20 &  56.00 &  25.90 & 84.40 & 36.00 & 40.30 \\
    DTG 5-shot \citeyearpar{li2023deliberate}   & 81.70 & 55.90 & 25.20 & 86.60 & 39.40 & 43.50
    \\
    BayLing \citeyearpar{zhang2023bayling} & \textbf{82.64} & 57.90 & 26.13 & 86.81 & \textbf{40.32} & \textbf{44.99} \\
    \hdashline
    GPT-3.5-turbo  & 80.60 & 58.40 & 26.93 & 81.48 & 37.80 & 42.85 \\
    \rowcolor{blue!5} \textbf{\hspace{3pt} + \method} & 80.47 & \textbf{58.50} & \textbf{27.04} & \textbf{88.06} & 39.86 & 44.40 \\
    \bottomrule
\end{tabular}
\caption{Experimental results on WMT22.}
\label{tab:mt}
\end{table*}

\paragraph{Case Study}
In Tables \ref{tab:case} and \ref{tab:case_nli}, we present case studies on paraphrasing and inference tasks. These demonstrate that \method not only excels in altering syntactic structures and employing a broader range of expressions, thereby enhancing the overall quality of paraphrasing, but also better captures sentence semantics.

\begin{table*}[t]
    \centering
    \vspace{-2mm}
    \begin{tabular}{p{0.9\linewidth}}
        \toprule
        \textbf{Source:} What can make Physics easy to learn? 
        \newline 
        \textbf{Target:} How can you make physics easy to learn? 
        \newline 
        \textbf{Vanilla Prediction:} \textcolor{orange}{What makes} learning Physics \textcolor{orange}{easier}? 
        \newline 
        \textbf{\method Prediction:}  \textcolor{cyan}{What factors} can help \textcolor{cyan}{simplify} the learning of Physics?
        \\
        \midrule
        \textbf{Source:} How do I create a new shell in a new terminal using C programming (Linux terminal)? 
        \newline 
        \textbf{Target:} How do I create a new Terminal and new shell in Linux using C programming? 
        \newline 
        \textbf{Vanilla Prediction:} \textcolor{orange}{How do I open} a new shell in a new terminal window using C programming \textcolor{orange}{in a Linux environment}? 
        \newline 
        \textbf{\method Prediction:} \textcolor{cyan}{What steps should I follow} to launch \textcolor{cyan}{a new shell in a separate terminal using C programming on a Linux system}?
        \\
        \bottomrule
    \end{tabular}
    \caption{Case study comparing \method and Vanilla prompt predictions for the paraphrasing task.}
    \label{tab:case}
    \vspace{-5mm}
\end{table*}

\begin{table*}[h]
    \centering
    \begin{tabular}{p{0.9\linewidth}}
         \toprule
         \textbf{Premise:} My article does not say or imply that real earnings growth only reflects retentions and that dividend growth must be zero or that all valuation techniques are out the window for firms that don't pay dividends.
         \newline
         \textbf{Hypothesis:} My article doesn't say or imply that real earnings growth reflects only retentions and that dividend growth must be zero or that valuation techniques are unused for firms which don't pay dividends, it implies that growth is a concrete and calculable statistic.
         \newline
         \textbf{Ground Truth:}  \textcolor{orange}{Neutral}
         \newline
         \textbf{Vanilla Prediction:}  \textcolor{cyan}{Contradict}
         \newline
         \textbf{\method Prediction:} The semantic parsing result of both sentences indicates that they are expressing the same idea, with some minor differences in wording. Therefore, the two sentences are  \textcolor{orange}{neutral} to each other.
         \newline
         \textbf{Analysis:}  In this case, SENSE correctly identifies that both sentences express the same idea with minor wording differences, thus labeling them as neutral. The baseline prediction incorrectly identifies a contradiction.
         \\
         \midrule
         \textbf{Premise:} Is there a favorite story or storybook or anything that you had when you were little that somebody read to you?
         \newline
         \textbf{Hypothesis:} You don't remember your childhood, do you?
         \newline
         \textbf{Ground Truth:}  \textcolor{orange}{Contradiction}
         \newline
         \textbf{Vanilla Prediction:} Sentence1  \textcolor{cyan}{entails} sentence2. If someone is asking about a favorite story or storybook from childhood, it implies that they believe the person has memories of being read to as a child.
         \newline
         \textbf{\method Prediction:} Based on the semantic parsing result, sentence1 is  \textcolor{cyan}{neutral} to sentence2. The first sentence is asking about a favorite story from childhood, while the second sentence is questioning the person's memory of their childhood. There is no direct contradiction or entailment between the two sentences. 
         \newline
         \textbf{Analysis:} SENSE detects that the first sentence is asking about a favorite story from childhood, while the second sentence is questioning the person's memory of their childhood. Since it regards the second sentence as a question without an answer, it believes that there is no direct contraction or entailment, but regardless, this example should not be the implied relationship as entailment predicted by the baseline.
         \\
         \bottomrule
    \end{tabular}
    \caption{Case study comparing \method and Vanilla prompt predictions for the Natural Language Inference task.}
    \label{tab:case_nli}
\end{table*}

\end{document}